\documentclass[letterpaper]{article} 
\usepackage{aaai24}  
\usepackage{times}  
\usepackage{helvet}  
\usepackage{courier}  
\usepackage[hyphens]{url}  
\usepackage{graphicx} 
\usepackage{natbib}  
\usepackage{caption} 
\usepackage{algorithm}
\usepackage{algorithmic}
\usepackage{mathrsfs}   
\usepackage{amsmath}
\usepackage{MnSymbol}
\usepackage{amsfonts}
\usepackage{color}
\usepackage{subfigure}
\usepackage{newfloat}
\usepackage{listings}
\DeclareCaptionStyle{ruled}{labelfont=normalfont,labelsep=colon,strut=off} 
\floatstyle{ruled}
\newfloat{listing}{tb}{lst}{}
\floatname{listing}{Listing}
\newtheorem{myDef}{Definition}  
\newtheorem{lemma}{Lemma}

\title{Situation-Dependent Causal Influence-Based Cooperative Multi-agent Reinforcement Learning}
\author{
    Xiao Du,
    Yutong Ye,
    Pengyu Zhang,
    Yaning Yang,
    Mingsong Chen,
    Ting Wang\thanks{Corresponding Author}
}
\affiliations{
Software Engineering Institute, East China Normal University\\

    \{52265902007, 52205902007, pengyu.zhang, 52215902011\}@stu.ecnu.edu.cn, \{mschen, twang\}@sei.ecnu.edu.cn
}

\usepackage{bibentry}

\begin{document}

\maketitle

\begin{abstract}
Learning to collaborate has witnessed significant progress in multi-agent reinforcement learning (MARL). However, promoting coordination among agents and enhancing exploration capabilities remain challenges. 
In multi-agent environments, interactions between agents are limited in specific situations. 
Effective collaboration between agents thus requires a nuanced understanding of when and how agents' actions influence others.
To this end, in this paper, we propose a novel MARL algorithm named Situation-Dependent Causal Influence-Based Cooperative Multi-agent Reinforcement Learning (SCIC), which
incorporates a novel Intrinsic reward mechanism based on a new cooperation criterion measured by situation-dependent causal influence among agents.
Our approach aims to detect inter-agent causal influences in specific situations based on the criterion using causal intervention and conditional mutual information. This effectively assists agents in exploring states that can positively impact other agents, thus promoting cooperation between agents.
The resulting update links coordinated exploration and intrinsic reward distribution, which enhance overall collaboration and performance.
Experimental results on various MARL benchmarks demonstrate the superiority of our method compared to state-of-the-art approaches.
\end{abstract}

\section{Introduction}\label{Introduction}
Along with the rapid development of deep reinforcement learning (RL), MARL has attracted increasing attention in recent years and witnessed significant in real-world problems such as traffic light control \cite{9103316},  coordination of autonomous vehicles \cite{2020Deep}, and robotics control \cite{10078380}, which can be effectively modeled as multi-agent game models. 
The most original MARL training method is the complete independent training of each agent, in which other agents are treated as part of the environment without any coordinated interaction between the behaviors of agents. 
Nonetheless, this training method suffers from the dilemma within a non-stationarity environment. 

The Centralized Training with Decentralized Execution (CTDE) emerges as a widely adopted solution for mitigating the challenges posed by non-stationary environments, where centralized training allows agents to access information about other agents. However, due to partial observability, during the execution phase, only local behavior-observation information can be relied upon.
Both policy-based and value-based approaches have been introduced for CTDE, such as MADDPG \cite{lowe2017multi}, COMA \cite{foerster2018counterfactual}, MAAC \cite{iqbal2019actor}, and QMIX \cite{2018QMIX}. 
Although global information can be accessed during the centralized training phase, these methods make an assumption that decentralized policies remain independent of each other such that the joint policy can be expressed as a product of independent policies. Thus, two pivotal challenges stem from the above issues. 
Firstly, the complete independence of agents' policies disregards the influence of other agents, thereby restricting the agents from learning coordinated behavior.
Secondly, optimizing decentralized strategies for multiple agents solely based on task-dependent dense reward signals often proves inefficient, particularly when these reward signals are stochastic or sparse.
Some studies \cite{2020Multi,2019Probabilistic} suggest that to alleviate the non-stationary issue and achieve collaboration between agents, it is imperative for agents to consider their impact on the behavior of other agents when making decisions.
Other works \cite{2019Signal,2019MAVEN,KimJCS23,LiTYHSZHTTW22} propose to quantify the correlation of agent behaviors through mutual information (MI), so as to maximize the correlation of agent behaviors to enhance collaboration.
MI has been identified as an effective intrinsic reward in promoting coordination in these works. 
Unfortunately, effectively coordinating the simultaneous actions of multi-agents in these approaches remains a challenging and intractable issue.
To this end, the primary objective of this study is to address this challenge from a causal inference perspective.  

In the context of control, if an agent can control other agents, it signifies that the agent can impact other agents through its actions. However, there is an underappreciated aspect hidden in this seemingly obvious observation, which is that the causal effects of behavior are state-dependent. 
Considering the multi-robot navigation scenario, when the current position of robot $a$ is close to robot $b$ and aligned with the current direction of movement of robot $b$, then robot $a$ evidently exerts a strong influence on robot $b$, irrespective of the next action of robot $a$. Conversely, if robot $a$ is situated far from robot $b$ and not in the movement direction of robot $b$, then robot $a$ has a comparatively weaker influence on robot $b$. 
Generally, there are situations in which agents have immediate causal influence, while in other situations, such influence is absent.
The key intuition of this study lies in the recognition that the states that enable an agent to have the ability to influence other agents of interest are important from both exploration and collaboration perspectives, and we name these states of importance as \textit{significant states}. 
Due to the fact that the initial states of an agent are rarely able to control the agents of interest, leading to inefficient training, these initial states are typically not regarded as \textit{significant states}.
Nonetheless, recognizing that the \textit{significant states} are conducive to inter-agent collaboration, it becomes imperative for the training process to prioritize and proactively explore these \textit{significant states}.
This exploration impels agents to consider the state-dependent causal effects among agents during the exploration process.

In this work, we exploit situation-dependent nature to measure the causal influence between agents. We propose an algorithm designed to achieve coordination among agents and coordinate agents' exploration, which gives agents an intrinsic reward based on state-dependent causal influence. Specifically, at each time step, all agents measure the causal influence between their actions in the current state and the next state of other agents, 
which is used to quantify the extent of causality between the agent and other agents at the current moment.
The mean of the causal influence originating from other agents is then taken as the agent's intrinsic reward. Furthermore, the computation of situation-dependent causal effects leverages conditional mutual information, which reliably identifies the \textit{significant states}. This intrinsic reward mechanism effectively facilitates the detection of \textit{significant states}, thereby enhancing the collaboration between agents.
It's worth noting that calculating mutual information of continuous variables is known to be challenging.
Compared to the variational information maximizing-based algorithms \cite{ChalkMT16, KolchinskyTW19}, MINE \cite{abs-1801-04062}, which learns a neural estimate of the mutual information, has shown superior performance \cite{HjelmFLGBTB19, VelickovicFHLBH19}.  Hence, we employ MINE to learn the conditional mutual information between agents by utilizing a forward dynamics model. To summarize, this paper makes the following three major contributions:
\begin{itemize}
    \item We model multi-agent reinforcement learning as a causal graph model by explicitly modeling the influence of causal factors in a multi-agent setting.
    \item We formalize the causal influence between agents as situation-dependent instead of action-dependent. Accordingly, a new Intrinsic Reward method with peer incentives is further proposed to promote the cooperation between agents using state-dependent causal influence, which is measured based on intervention and conditional mutual information.
    \item We conduct comprehensive experiments on various MARL benchmarks. The experimental results demonstrate that our approach outperforms other competitive methods, and the learned intrinsic reward proves to be conducive to learning better policies that achieve agents' better cooperation in these complex tasks.
    
\end{itemize}

\section{Related Work}\label{Related_work}
As an important training paradigm for solving agent coordination problems, CTDE has been widely adopted in MARL in recent years, where VDN \cite{SunehagLGCZJLSL18} and QMIX based on value function decomposition, and MADDPG and MAAC based on centralized critic are typical CTDE-based algorithms. Since our approach is designed to utilize causal influence as an intrinsic reward to enhance inter-agent collaboration in MARL, in this section we will briefly review related work in terms of Intrinsic Reward for MARL and causality in reinforcement learning.

\subsection{Intrinsic Reward for MARL}
Intrinsic rewards help agents learn useful policies across a wide variety of tasks and environments, even sometimes with sparse environmental rewards. At each time step, the agents receive not only environmental rewards but also intrinsic rewards. Intrinsic rewards can help agents improve their exploration ability and enhance social influence. In reinforcement learning, all kinds of information have various signals as intrinsic rewards, such as empowerment \cite{MohamedR15}, model surprise \cite{BlaesPZM19}, information gain \cite{HouthooftCCDSTA16}, and learning progress \cite{BlaesPZM19}. 
In MARL, some existing approaches utilize the correlation or influence of agents as intrinsic rewards to facilitate collaboration. EITI \cite{0001WWZ20} leverages MI to capture the influence between an agent's current trajectory and the next states of other agents in the environment, which is used as an intrinsic reward to encourage agents to explore cooperatively. SI \cite{JaquesLHGOSLF19} proposes a method that utilizes social influence as an intrinsic reward, measured by the MI between an agent's current action and the estimated next action of other agents, to achieve coordination and communication in MARL. VM3-AC \cite{KimJCS23} is similar to the idea of coordinating inter-agent behavior in SI-MOA, except that additional latent variables are introduced to induce nonzero mutual information between multi-agent actions, and the policy iteration algorithm is modified based on MI. PMIC \cite{LiTYHSZHTTW22} utilizes the MI between global states and joint actions as a new standard. Based on this standard, it maximizes the MI related to behaviors that are conducive to collaboration and minimizes the MI related to behaviors that are not conducive to collaboration, breaking the current suboptimal collaboration and learning higher-level collaborative behaviors.
Comparatively, our approach tackles the challenge of collaboration among agents from a causal inference perspective, which detects situation-dependent causal relationships among agents as intrinsic rewards to facilitate the exploration of inter-agent coordination.

\subsection{Causality in Reinforcement Learning}
The combination of reinforcement learning and causality has achieved some progress. The work \cite{abs-2102-11107} utilizes causal modeling to achieve better state abstraction, enabling the agent to concentrate on key aspects and indirectly improving sampling efficiency. The work \cite{SeitzerSM21} proposes integrating measures of causal influence into reinforcement learning algorithms to address the problems of exploration and learning in the robot manipulation environment. The work \cite{LeeZSGK21} utilizes causal intervention to identify the most relevant state variables for completing a task, thereby reducing the dimensionality of the state space. 
The work \cite{PitisCG20} utilizes influence detection to create counterfactual data to enhance the training of RL agents. 
\cite{MAES2007274} extends causal Bayesian networks to multi-agent models, which inspires us to develop a causal graph model for MARL.

\section{Background}\label{Background}

\subsection{Preliminaries}
In this work, we consider the fully cooperative multi-agent game in the partially observable setting, which can be modeled as a decentralized partially observable Markov Decision Process (Dec-POMDP) \cite{OliehoekA16}. The Dec-POMDP is formally defined as a tuple $\langle\mathcal{I,S,U,O,P,R,\gamma}\rangle$, where $\mathcal I$ = $\lbrace 1,2,....N \rbrace$ denotes the finite set of agents, $ s_t 
 \in \mathcal S$ denotes the set of joint states that cannot be observed by agents at the time step $t$. At each time step $t$, each agent $i$ $\in \mathcal{I}$ can only observe its local observation $o_t^i$ from observation function $\mathcal{O}(s_t, i)$ and chooses an action $u_t^i \in \mathcal{U}^i$ according to its policy $\pi(\dot|o_t^i)$, forming a joint action $u_t \in \mathcal{U}$. After executing $u_t$ in environment, each agent $i$ achieves a shared extrinsic reward $r_t$ from the reward function $\mathcal{R}(s_t, u_t)$ with a discount factor $\gamma \in [$0,1$)$ and the next state $s_{t+1}$ according to the transition function $\mathcal{P}(s_{t+1}|s_t,u_t)$. The goal of a collaborative team is to find a joint policy $\pi^*$ that can maximize the expected extrinsic discount return ${E}[\sum_{k=1}^\infty \gamma^t r_t]$ under the setting of fully-cooperative MARL.

In our approach, we adopt a CTDE paradigm, which has been a widely considered training paradigm in recent efforts in MARL \cite{SunehagLGCZJLSL18, 2018QMIX, lowe2017multi,iqbal2019actor}. During training, each agent can access to full information including the states, actions, rewards, and actions of other agents, while decentralized execution is conditioned solely on individual observation $o_t^i$.

\subsection{Causal Graphical Models}
A causal graphical model (CGM) \cite{Shanmugam01} is commonly represented as directed cyclic graphs (DAGs)  $\mathcal{G}= (V,E)$  and is defined by a distribution $\mathcal{P}$ over the set of random variables $\mathcal{X}$. The graph $\mathcal{G}$ consists of nodes $V$ and edges $\mathcal{E} \in V^2$ with $(v,v)$ for any $v \in V$. Each node $v_i$ is associated with a random variable $\mathcal{X}^i$ and each edge $(v_i \rightarrow v_j)$ represents that $\mathcal{X}^i$ is a direct cause of $\mathcal{X}^j$, i.e., $\mathcal{X}^i$ is called a parent of $\mathcal{X}^j$. The set of parents of $\mathcal{X}^j$ is denoted by $\mathbf{PA}_j^\mathcal{G}$. The distribution $\mathcal{P}$ can be represented as 
\begin{align}
p(X^1,...,X^v) = \prod \limits_{i=1}^v p_i(X^i|\mathbf{PA}_j^\mathcal{G}),
\end{align}
where $\mathbf{PA}_j^\mathcal{G} \subset \left\{X^1,...,X^v \right\} \backslash \left\{X^j \right\}$ denotes the set of parents of $X^j$. The CGM models the structure of the causality. To reveal the causal structure from the data distribution, we assume that CGM satisfies the Markov condition and the faithfulness assumption, which makes the independence consistent between the joint distribution $P(X^1,...,X^v)$ and the graph $\mathcal{G}$.

\subsection{Intervention}
Intervention sampling is a typical operation in causal discovery. Different from standard sampling, it sets the distribution of variables that require intervention in the causal graph model to a uniform distribution or fixed value, and then conducts sampling to obtain intervention data. Intervention data has more causal information than observational data, which is beneficial to measure causal relationships between random variables. In this work, we need to measure the influence of the current time step's action of an agent on the state of other agents in the next time step, which can be obtained by intervening in the current behavior distribution.

\section{Our Approach}\label{Approach}
\begin{figure*}[htb] \centering 
	\centering  
	\includegraphics[width=0.73\linewidth]{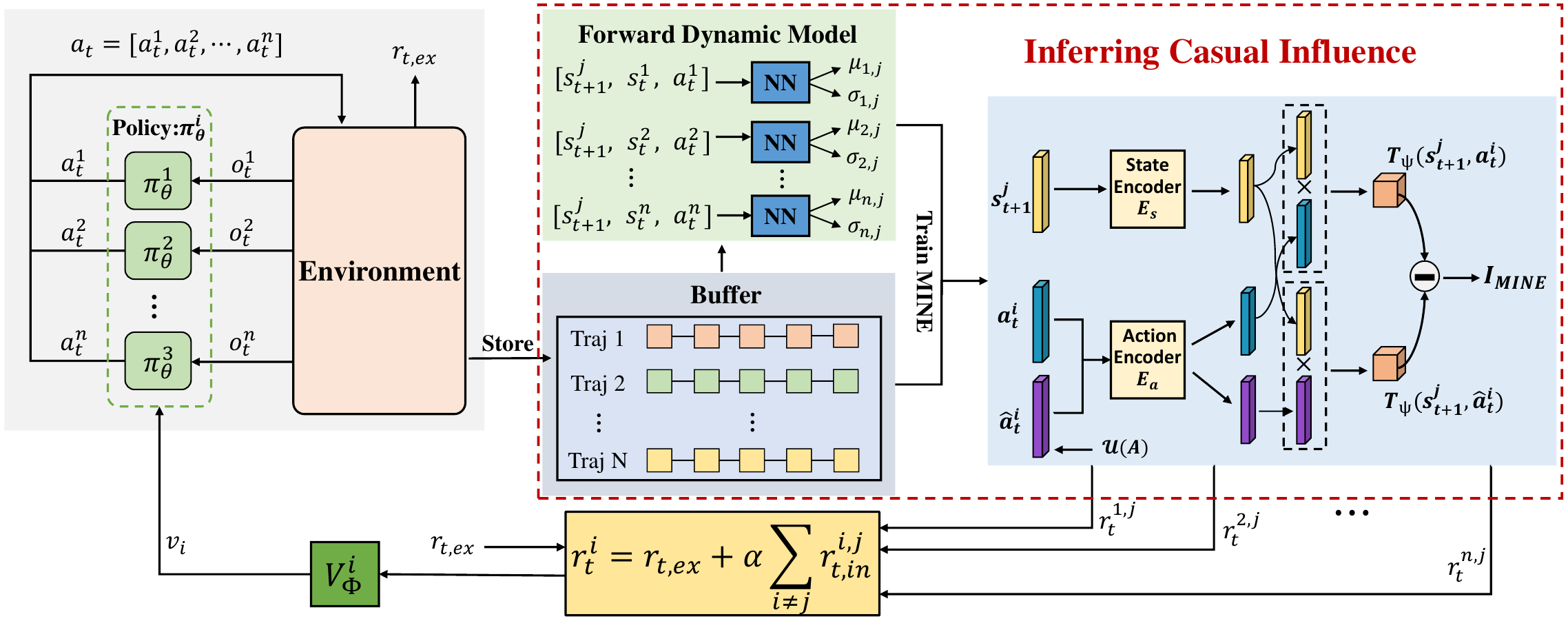}  
	\caption{An overview of SCIC framework.} \label{framework}
\end{figure*}

In this section, we present the design of our SCIC approach, in which the agents simultaneously learn a policy and an intrinsic reward function by maximizing the causal influence between agents, as illustrated in Figure \ref{framework}. Our SCIC approach detects the inter-agent causal influences in particular situations based on causal intervention and conditional mutual information, facilitating agents to explore states that can affect other agents, thereby promoting cooperation among agents. 

\subsection{Causal Influence-based  Intrinsic Reward Design}
A key component of SCIC is the intrinsic reward mechanism that brings each agent to a state where it can influence interested agents as much as possible. Intuitively, an agent is more likely to influence interested agents in certain states, and making agents reach these states as much as possible is more likely to enhance the causal influence between agents. We represent agent $j$ being able to be influenced by agent $i$ as ``agent $i$ being able to take control of agent $j$". Through mutual incentives between agents, agents can achieve friendly interaction and efficient cooperation with their peers. Thus, the causal influence of other agents on an agent can be viewed as a necessity for learning intrinsic reward mechanisms. We detect the causal influences between agents in particular situations using causal intervention and conditional mutual information. It is worth noting that the computation of causal influence is only executed during intensive training, where each agent can know the policies and actions of other agents.

\subsubsection{Multi-agent Causal Graph Model}

\begin{figure}[htb]  
	\centering  
	\includegraphics[width=0.67\linewidth]{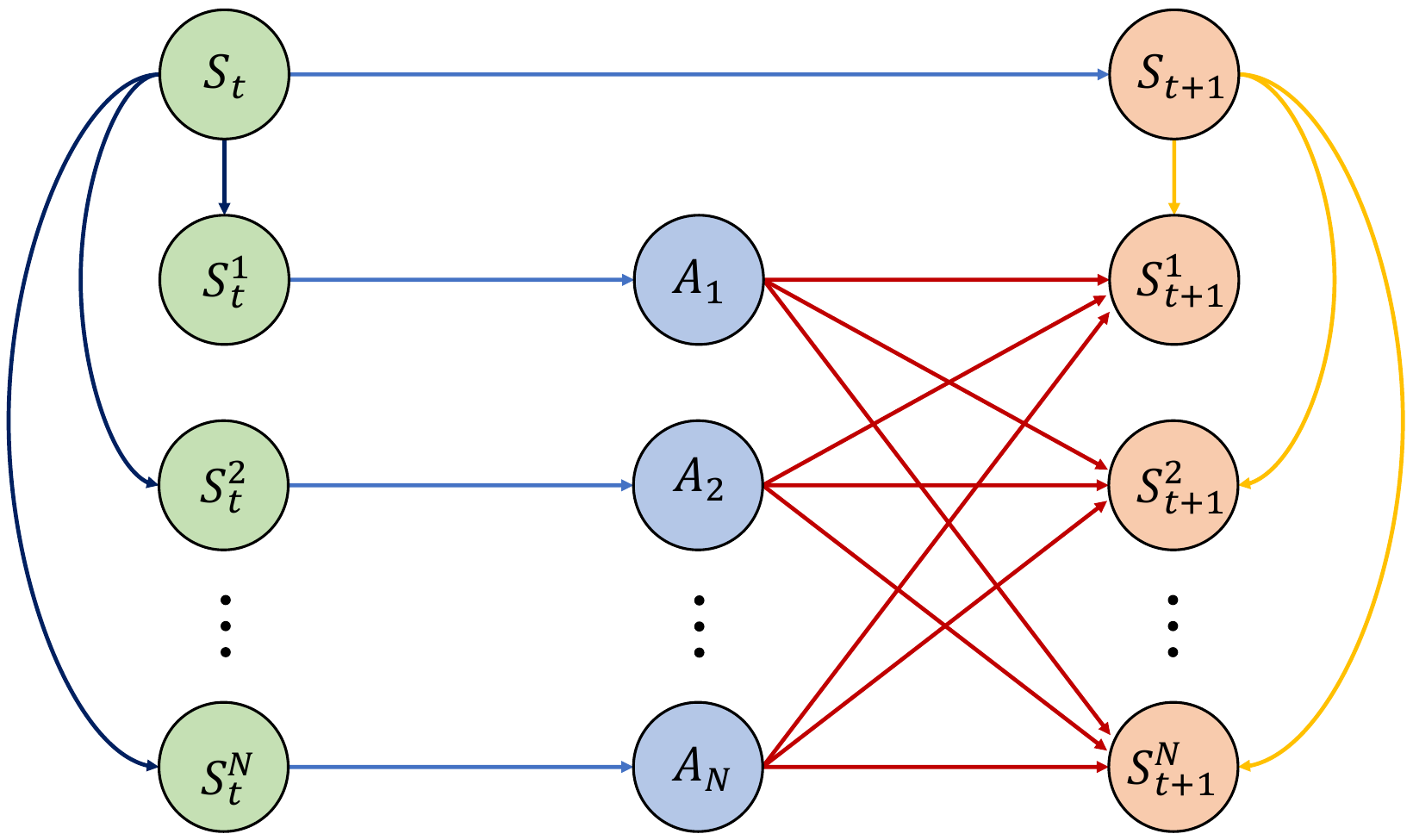}  
 	\caption{Multi-agent Causal Graph Model.
	}  \label{multi_agent_causal_graph}
\end{figure}

We extend the CGM to the case of decentralized multi-agents as Multi-agent Causal Graph Model (MACGM), where agents share an environment and have access to private and/or public variables of interest during centralized training. The one-step transition dynamics of MACGM at time step $t$  is modeled a causal graphical $\mathcal{G}$ (see Figure \ref{multi_agent_causal_graph}) over the set of random variables $\mathcal{V} = \left\{S_t,S_{t+1},S_t^1,...,S_t^N,A_t^1,...,A_t^N,S_{t+1}^1,...,S_{t+1}^N  \right\}$, consisting of a conditional distribution $P(V_i | PA(V_i))$, where $V_i$ represents an agent's state component e.g. $S_t$, or the agent's action component e.g. $A_t^1$. Apart from the actions computed by the policy $\pi(A_t|S_t)$, within a time step, there are no edges, i.e. no transient effects.
In the MACGM,  the actions $A_t$ at the $t$ time step are intended to affect $S_{t+1}$ by influencing the environment. 
To better express the causal relationship between actions and states, considering that the influence of $A_t$ on $S_{t+1}$ is actually the influence on the environment at time $t$+1, thus the influence on the environment can be differentiated into the influence on the state $S_{t+1}^j$ of each agent $j$ at $t$+1 time step. Nevertheless, at most time steps, there should be no instantaneous effects between agents in the world. In particular, an agent's sphere of influence is limited, i.e., its action $A$ can only affect other agents sparsely, which rests on two basic assumptions about the causal structure of the world. First, the multi-agent environment is composed of independent agents, according to the principle of independent causal mechanism (ICM) \cite{2018Elements}, which states that the generative process of the world is composed of autonomous modules. The second assumption is that latent influences between entities are spatially local and temporally sparse. We can view this as explaining the sparsity mechanism shift hypothesis, which suggests that natural distribution shift will be caused by changes in local mechanisms \cite{abs-2102-11107}. This can usually be traced back to the ICM principle, which states that intervention in one mechanism does not affect other mechanisms \cite{abs-1712-00961}. We believe that this is also due to the limited scope of intelligent agent intervention, which limits the breadth and frequency of mechanism changes. Therefore, in this work, we are interested in inferring the influence of the action of agent $i$ over other agents in a particular situation, i.e. a local inter-agent causal model. Next, we provide the following definitions:
\begin{myDef}
(Controllable State Variable) If the edge $A_t^i \rightarrow S_{t+1}^j$ in the graph is ``active", $S_{t+1}^j$ is a controllable state variable of $A_t^i$.
\end{myDef}

\begin{myDef}
(Uncontrollable State Variable) If the edge $A_t^i \rightarrow S_{t+1}^j$ in the graph is ``inactive", $S_{t+1}^j$ is a uncontrollable state variable of $A_t^i$.
\end{myDef}
Given these definitions, in this work, our aim is to detect whether the state is a ``Controllable State Variable" for other agents in a particular situation, i.e. whether the presence of red arrows in Figure \ref{multi_agent_causal_graph} is ``active". 

\subsubsection{The Cause of an Influence}
When is action $A=a$ the cause of outcome $S=s$? Inspired by the ``but-for" test, i.e. ``Without $A = a$, $S = s$ would not have happened.", we can derive that $A = a$ is a necessary condition for $S = s$ to happen, and when $A$ changes, $S$ also gets a different value. This fits with the algorithmic view of causality: if the value of $S$ is determined by the value of $A$, then $A$ is the cause of $S$. The ``but-for" test yields potentially counter-intuitive assessments. Considering the influence between two agents located close to each other, the behavior of agent $a$ is regarded as the cause of the influence on another agent $b$, because different behaviors of agent $a$ will lead to different behavior choices of agent $b$, thus affecting the next state of $b$. Algorithmically, the behavior of the agent $a$ needs to be known in order to determine the effect on agent $b$ - all possible behaviors of agent $a$ are considered as causes. This means that we cannot distinguish whether the agent's behavior is the cause, but only whether the agent has a causal influence on other agents in the current situation.
\subsubsection{Intervention for Causal Inference}
As discussed above, the causal relationship between agents depends on the current situation rather than the behavior chosen by the agents. In order to detect whether there is a causal relationship between agents, we utilize an intervention method to achieve it.
Formally, we define ``agent $i$ can causally affect agent $j$ in the current situation" (or ``agent $i$ takes control of agent $j$") if there exists an edge $A_t^i \rightarrow S_{t+1}^j$ in causal graph (as shown in Figure \ref{multi_agent_causal_graph}) under all interventions $do$$(A_t^i$=$\pi(a_t^i|s_t^i))$ with $\pi$ having full support. 
According to Markov property and the faithfulness assumption of the causal graph model, if $A_t^i \nupmodels S_{t+1}^j | S_t^i=s_t^i$, then there must be an unblocked edge from $A_t^i$ to $S_{t+1}^j$ in a causal graph $\mathcal{G}$. Since the path over $S_{t+1}^j$ is blocked by observing $S_t^i$, while assuming no instantaneous influences, the direct path $A_t^i \rightarrow S_{t+1}^j$  is the only possible path.
Therefore, in causal graph $\mathcal{G}$, there is an edge $A_t^i \rightarrow  S_{t+1}^j$ under the intervention $do($$A_t^i$=$\pi(a_t^i|s_t^i))$ if $A_t^i \nupmodels S_{t+1}^j | S_t^i$. 
The following lemma shows that a conclusion taken from an intervention generalize to numerous interventions with $\pi$ having full support (proofs in Suppl. A). According to the lemma, the conclusion from an intervention can determine whether there is a causal influence between agents.
\begin{lemma}\label{lemma1}
If $A_t^i \nupmodels S_{t+1}^j$ holds under an intervention $do($$A_t^i:=\pi(a_t^i|s_t^i))$, then the dependence holds and the edge $A_t^i \rightarrow S_{t+1}^j$ exits  under all interventions with $\pi$ having full support. If $A_t^i \upmodels S_{t+1}^j$ holds under an intervention $do($$A_t^i:=\pi(a_t^i|s_t^i))$ with $\pi$ having full support, then the independence holds and the edge $A_t^i \rightarrow S_{t+1}^j$ does not exit under all interventions. 
\end{lemma}
\subsubsection{Causal Influence Detection between Agents}
Our goal is to measure state-dependent causal influence between agents, which is linked to the independence $A_t^i \upmodels S_{t+1}^j | S_t^i=s_t^i$ or dependence $A_t^i \nupmodels S_{t+1}^j | S_t^i=s_t^i$. Conditional mutual information (CMI) is a well-known measure of dependence, which is proposed to be utilized as a measure of causal influence (CI)  between agents. 
If CMI \textgreater 0, it suggests that $A_t^i$ is necessary to predict $S_{t+1}^j$, $A_t^i \nupmodels S_{t+1}^j | S_t^i=s_t^i$ is true, and the causal path $A_t^i \rightarrow S_{t+1}^j$ exists. However, mutual information (MI) of continuous variables is notoriously difficult to compute in real-world settings. Compared to the variational inference-based approaches, MINE-based algorithms have shown superior performance. Motivated by MINE, our approach learns a neural estimate of MI, which utilizes a lower bound to approximate the MI. 

\begin{align}
 &CI^{ij} := I(S_{t+1}^j; A_{t}^i | S_{t}^i)   
 \\&=  KL\Big(P_{S_{t+1}^j,A_{t}^i|s_t^i} \big\Vert P_{S_{t+1}^j|s_t^i} \otimes P_{A_{t}^i|s_t^i}\Big)  \label{MINE2} 
 \\&= \mathop{sup}\limits_{T:\Omega \rightarrow R} \mathbb{E}_{p(S_{t+1}^j,A_t^i | s_t^i)}[T] -log(\mathbb{E}_{p(S_{t+1}^j | s_t^i)p(A_t^i | s_t^i)}[e^T]) 
 \\&\geq \mathop{sup}\limits_{\psi \in \Psi} \mathbb{E}_{p(S_{t+1}^j,A_t^i | s_t^i)}[T_\psi] -log(\mathbb{E}_{p(S_{t+1}^j | s_t^i)p(A_t^i | s_t^i)}[e^{T_\psi}]). \label{MINE3} 
\end{align}

First, the CMI formulation is rewritten as Equation (\ref{MINE2}) using the Donsker-Varadhan representation \cite{2010Asympototic}. The input space $\Omega$ is a domain of $\mathbb{R}_d$. The upper bound holds for all functions $T$ such that two expectations are finite.  Then, the CMI in the Donsker-Varadhan representation is derived with a lower bound using the compression lemma in the PAC-Bayes literature in Equation (\ref{MINE3}) \cite{2006On}. The statistical model $T$ is parameterized by a deep neural network with parameter $\psi$. 
In addition, since the data in the off-policy reinforcement learning algorithm stems from a mixture of different policies, the agent's sampling strategy cannot be used for intervention. Fortunately, Lemma \ref{lemma1} has shown that a single policy is sufficient to demonstrate (in-)dependence. Therefore, we select a uniform distribution $\mathcal{U}(\mathcal{A})$ over the action space as the intervention policy.

One thing to note is that the forward dynamics model $p(s_{t+1}^j|s_t^i,a_t^i)$ and the distribution $p(s_{t+1}^j|s_t^i)$ need to be calculated. The following will elaborate on how to calculate these two distributions. Computing these two distributions involves representing complex distributions, computing high-dimensional integrals, and only limited data. In addition, each state $s_t^i$ actually can only be seen once in continuous space. 
Since methods based on non-parametric estimation do not scale well to high dimensions, we address this issue by learning neural network models with appropriately simplifying assumptions. As shown in Figure \ref{framework}, we first utilize the sampled data from the buffer to estimate the transition distribution $p(s_{t+1}^j|s_t^i,a_t^i)$, assuming that the transition distribution is a normal distribution. Then, the forward dynamics model is utilized to compute the transition marginal distribution $p(s_{t+1}^j|s_t^i)$ by marginalizing out the actions $p(s_{t+1}^j|s_t^i) = \int \pi(a_t^i|s_t^i)p(s_{t+1}^j,a_t^i|s_t^i)$. Actually, we utilize Monte-Carlo to approximate the mixture $p(s_{t+1}^j|s_t^i) \approx \frac{1}{K} \sum_{k=1}^K p(s_{t+1}^j,a_t^i|s_t^{i,(k)})$, instead of integrals. During the training process, the dynamic model $p(s_{t+1}^j|s_t^i)$ is trained simultaneously with the statistical network $T$ and agents' policy network.

\subsection{Training with Causal Influence as Intrinsic Reward}
With the causal influence estimations introduced in the previous subsection, the goal of the following is to learn the joint policy that maximizes the expected discounted reward, using causal influence between agents as an intrinsic reward. Specifically, agent $i$ receives a joint reward, combining the extrinsic team reward and the intrinsic reward from peer causal influence, that can be represented as 
\begin{align}
r_{t,total}^i = r_{t,ex} + \alpha \sum \limits_{j \neq i} CI^{i,j},
\end{align}
where $CI^{i,j}$ represents the causal influence of $A_i^t$ on $S_{t+1}^j$. $\alpha$ is a hyper-parameter, which is utilized to balance the intrinsic reward and the extrinsic reward. The intrinsic reward $r_{t,in}^i$ of agent $i$ is represented as $r_{t,in}^i = \sum_{j\neq i} CI^{i,j}$. In particular, each agent needs to learn a policy to maximize the conventional objective $J(\pi): J(\pi_\theta^i) =  \mathbb{E}\lbrack \sum \limits_{t=0}^\infty \gamma^t r_{t,total}^i\rbrack$. By maximizing the conventional objective, agent $i$ can take control of other interested agents. In principle, our proposed intrinsic reward mechanism can be combined with different CTDE-based MARL algorithms. In this work, our method is built upon MADDPG. The joint action-value function of each agent $i$ is approximated by the joint action-value network $\mathcal{Q}_\phi$ by minimizing the following loss
\begin{align} \label{Q_Value}
\mathcal{L_{\mathcal{Q}}}(\phi_i) = \mathbb{E}_{s_t,a_t,r_t,s_{t+1} \sim 
D} \lbrack (\hat{y}_i - \mathcal{Q}_{\phi_i}(s_t,a_t))^{2} \rbrack,
\end{align}
where $\hat{y} = r_t + r_{t,in}^i + \gamma \mathcal{Q}_{\phi_i^{'}}(s_{t+1},\pi_{\theta^{'}}(s_{t+1}))$, $\phi_i$ indicates the parameters of critic network, and $\phi_i^{'}$ and $\theta^{'}$ are the parameters of target networks.
Then, the policy of agent $i$ is updated by minimizing the loss
\begin{align}\label{Actor}
\mathcal{L_{\pi}}(\theta_i) = \mathbb{E}_{s_t\sim \mathcal{D}} \lbrack -\mathcal{Q}_{\phi_i}(s_t,\pi_{\theta}(\cdot|s_t))  \rbrack,
\end{align}
where $\theta = (\theta_1,...,\theta_n)$ are the parameters of actor networks.
The training algorithm is described in Algorithm \ref{alg:algorithm}.

\begin{algorithm}[tb]
\caption{Training algorithm}
\label{alg:algorithm}
\textbf{Initialize}: The critic networks $\phi = (\phi_1,...,\phi_n)$, the actor networks $\theta=(\theta_1,...,\theta_n)$, experience replay buffer $\mathcal{D}$ and  target networks $\phi^{'}$, $\theta^{'}$, the parameters of the statistic networks $\mathcal{T}$, the forward dynamic models f = $\lbrace f_i \rbrace_{i=1}^n$ and the state encoder networks e.\\  
\begin{algorithmic}[1] 
\WHILE{episode \textless $M$ }
\FOR{t = (1,...,T)}
\STATE Collecting $s_{t+1}$=$\{s_{t+1}^i\}_{i=1}^n$ and extrinsic reward $r_{t,ex}$ by executing joint actions $a_t$ via collecting $a_t^i\sim\pi_{\theta_i}(s_t^i)$. \\
\ENDFOR
\STATE Store episode trajectory $\lbrace$$s_t$, $a_t$,$s_{t+1}$,$r_{t,ex}$$\rbrace$ from the multi-agent environment to replay buffer $\mathcal{D}$.
\STATE Sample a batch data of transition form buffer $\mathcal{D}$.
\STATE Update forward dynamic model for each agent $i$. 
\STATE Update $\mathcal{T}$  via Equation \ref{MINE3}. 
\STATE Compute intrinsic reward $r_{t,in}^i$ for each agent $i$ via $r_{t,in}^i = \sum_{j\neq i} CI^{i,j}$.
\STATE Update critic networks via Equation \ref {Q_Value} and intrinsic reward.
\STATE Update actor networks via Equation \ref {Actor}.
\ENDWHILE \\
\end{algorithmic}
\end{algorithm}

\section{Performance Evaluation}\label{Evaluation}
To verify the effectiveness of our proposed method, in this section, we conduct extensive experiments to evaluate SCIC on various multi-agent tasks and compare SCIC with state-of-the-art methods. Moreover, we also evaluate the effectiveness of the components of the proposed method by ablation experiments.
\subsection{Multi-Agent Task Benchmark}
We evaluate our proposed approach on three benchmark multi-agent tasks: Partial Observation Cooperative Predator Prey, Cooperative Navigation, and Cooperative Line Control. The benchmarks' environment is implemented in a Multi-Agent Particle Environment (\cite{lowe2017multi}), where agents can follow a dual integrator dynamic model to move in 2D space. Like Cooperative Navigation, Cooperative Predator Prey is a well-known evaluation task for MARL. The Cooperative Line environment is a complex task environment, where there are $M$ agents and 2 targets, with the goal of allowing agents to evenly distribute themselves on the line between two targets.  
All algorithms are trained in a Linux server with a 2.30 GHz Xeon(R) CPU and two Nvidia 4090 graphics cards. The learning rates of the critic network and the actor network are set to 0.001. The discount factor $\gamma$ is set to 0.95. Each episode lasts up to 25 timesteps. To estimate the transition marginal distribution $p(s_{t+1}^j|s_t^i)$, the number $K$ of per Monte-Carlo sample is set to 64.

\subsection{Baselines}
To verify the superiority of our method, we employed three baseline algorithms for comparison, namely PMIC, MADDPG and SI, with a relaxation that all SI agents are
equipped with the social influence reward. Among them, PMIC and SI are currently state-of-the-art MARL algorithms using intrinsic rewards in Multi-Agent Particle Environment (MPE). Since our proposed algorithm is based on MADDPG, MADDPG is also utilized as a baseline algorithm for comparison. Additionally, to verify the effectiveness of the components of the proposed method, some ablation studies also will be provided. Furthermore, considering the significant role played by the temperature parameters of intrinsic reward in balancing the relative importance between extrinsic and intrinsic reward, we also provide an experimental study on the temperature parameters.

\subsection{Experimental Results} 
\begin{figure}[ht] \centering
	\subfigure[Predator Prey (3 agents).] {
	\centering
	\label{pre3}
	\includegraphics[width=0.47\linewidth]{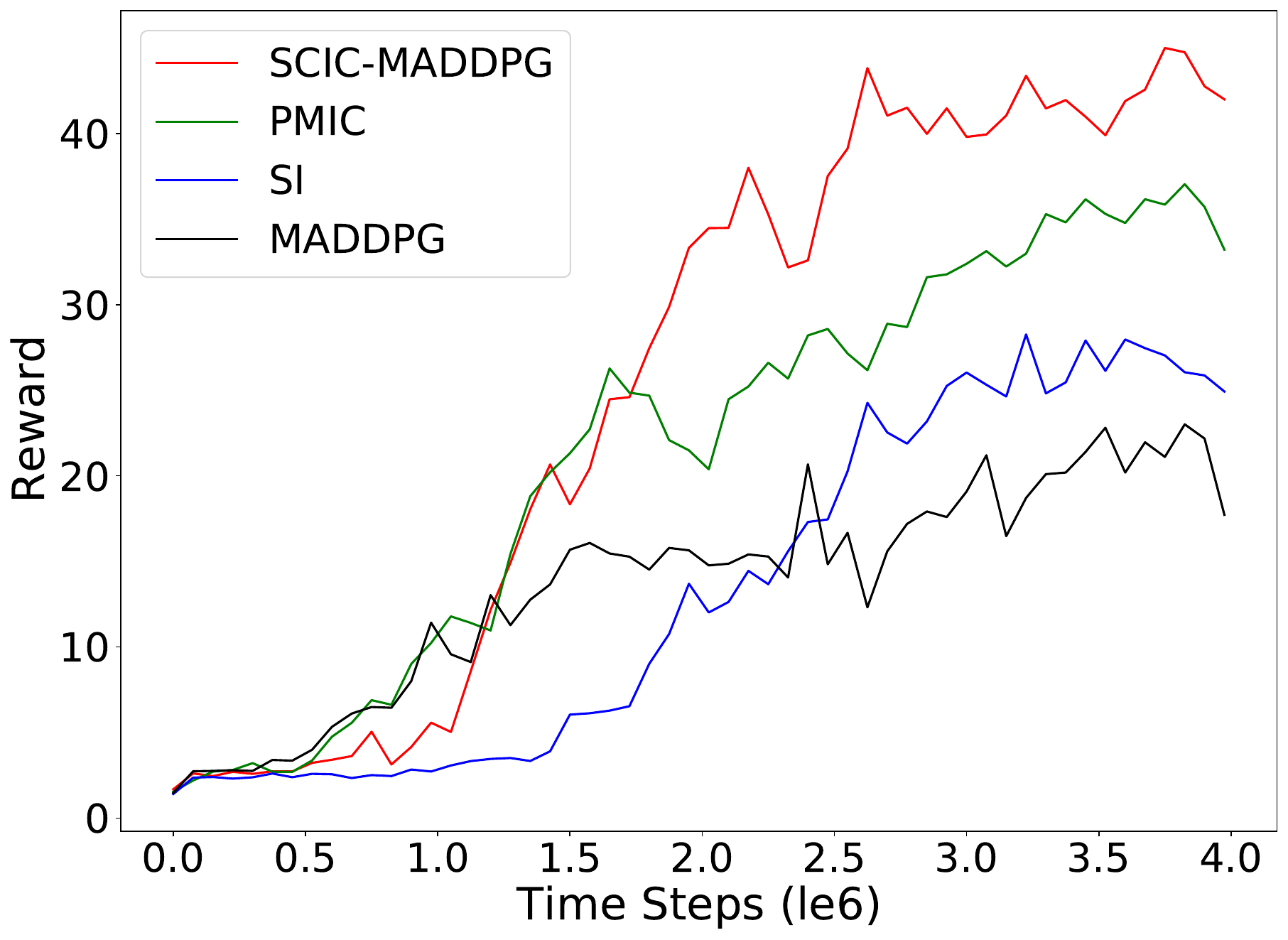}
	}
	\subfigure[Predator Prey (4 agents)] {
	\centering
	\label{pre4}
	\includegraphics[width=0.47\linewidth]{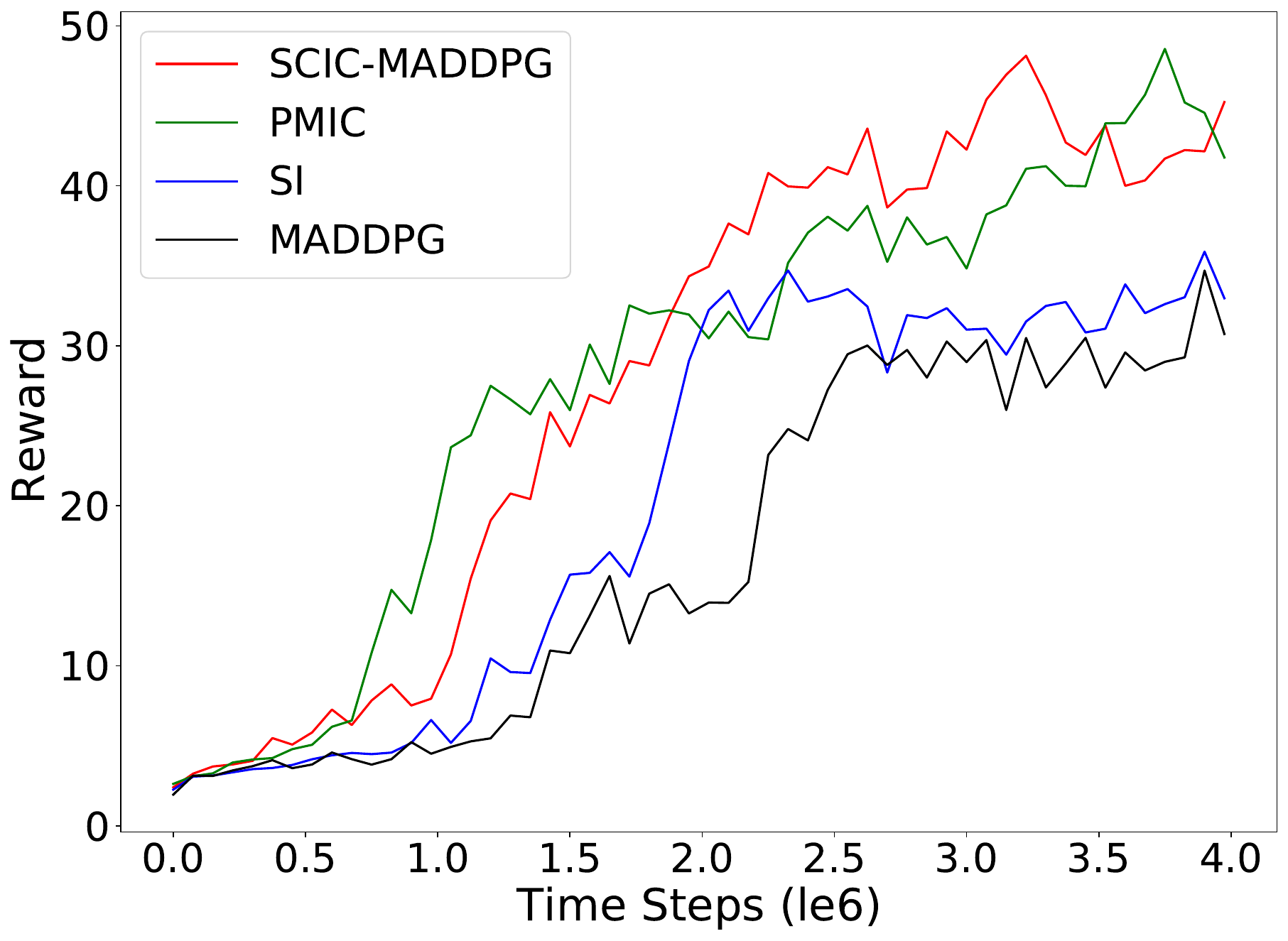}
	}
	\subfigure[Predator Prey (5 agents).] {
	\centering
	\label{pre5}
	\includegraphics[width=0.46\linewidth]{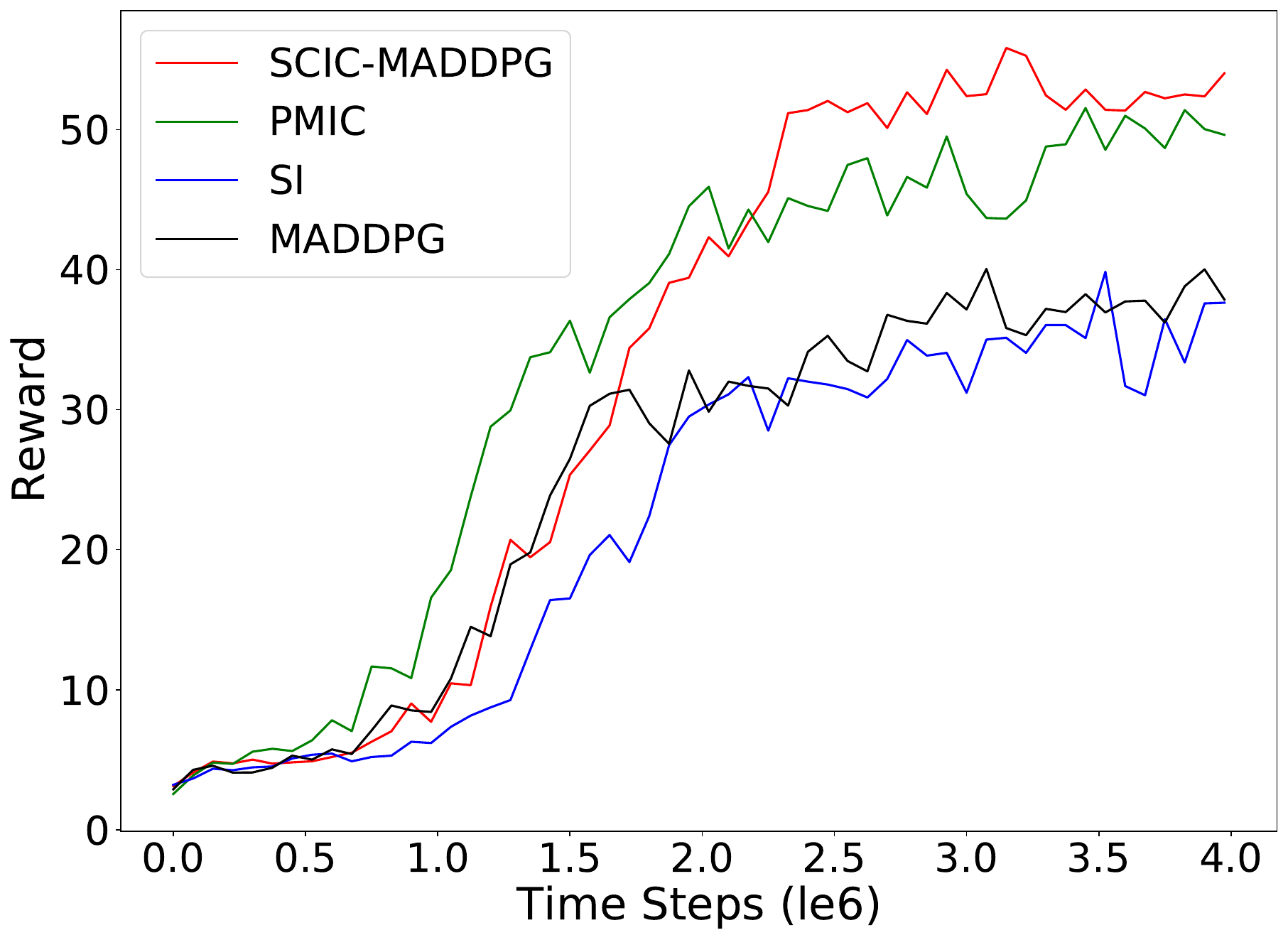}
	}
  \hspace{-1mm}
        \subfigure[Cooperative Navi. (3 agents).] {
	\centering
	\label{navigation3}
	\includegraphics[width=0.48\linewidth]{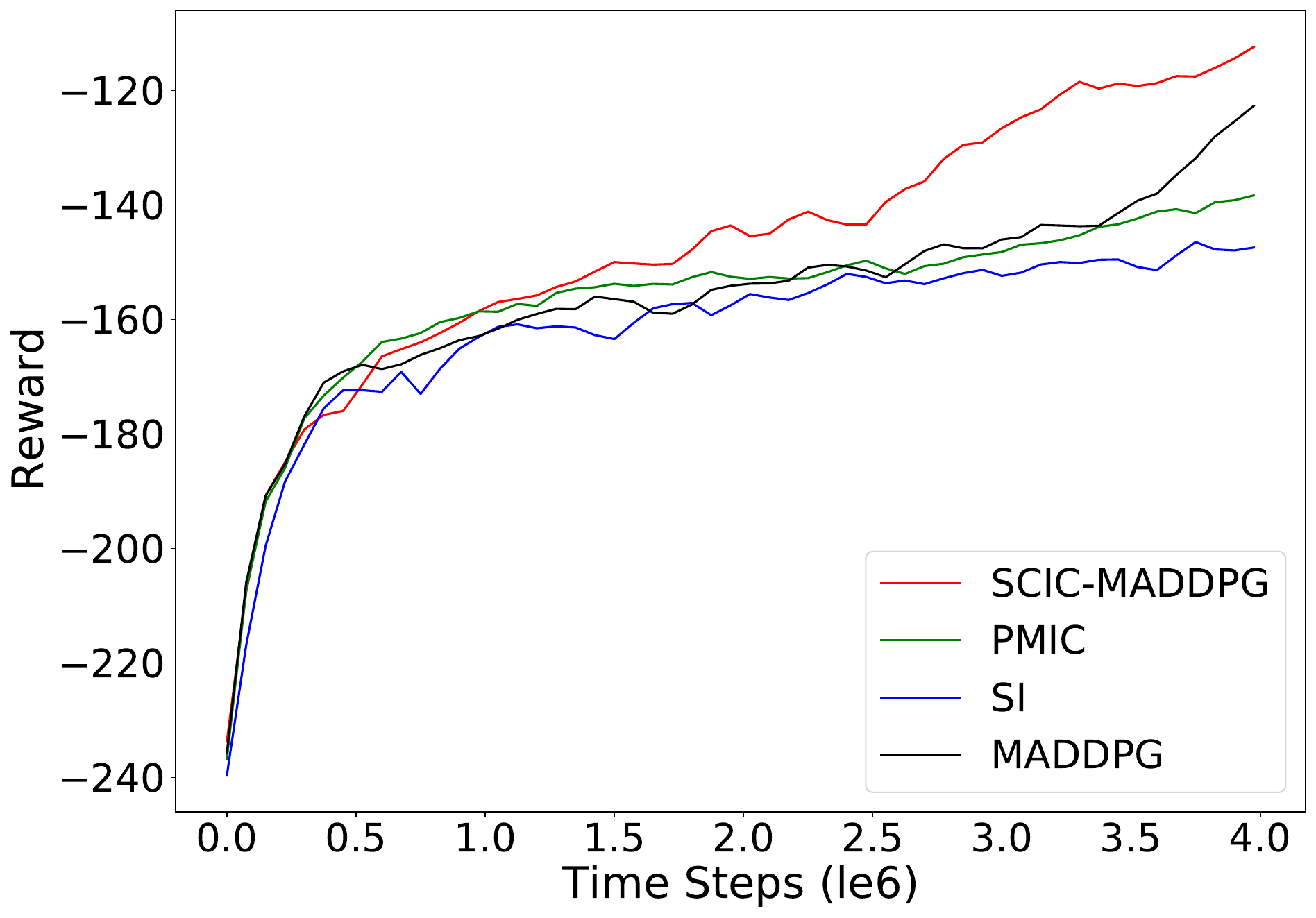}
	}
	\subfigure[Cooperative Navi. (4 agents).] {
	\centering
	\label{navigation4}
	\includegraphics[width=0.47\linewidth]{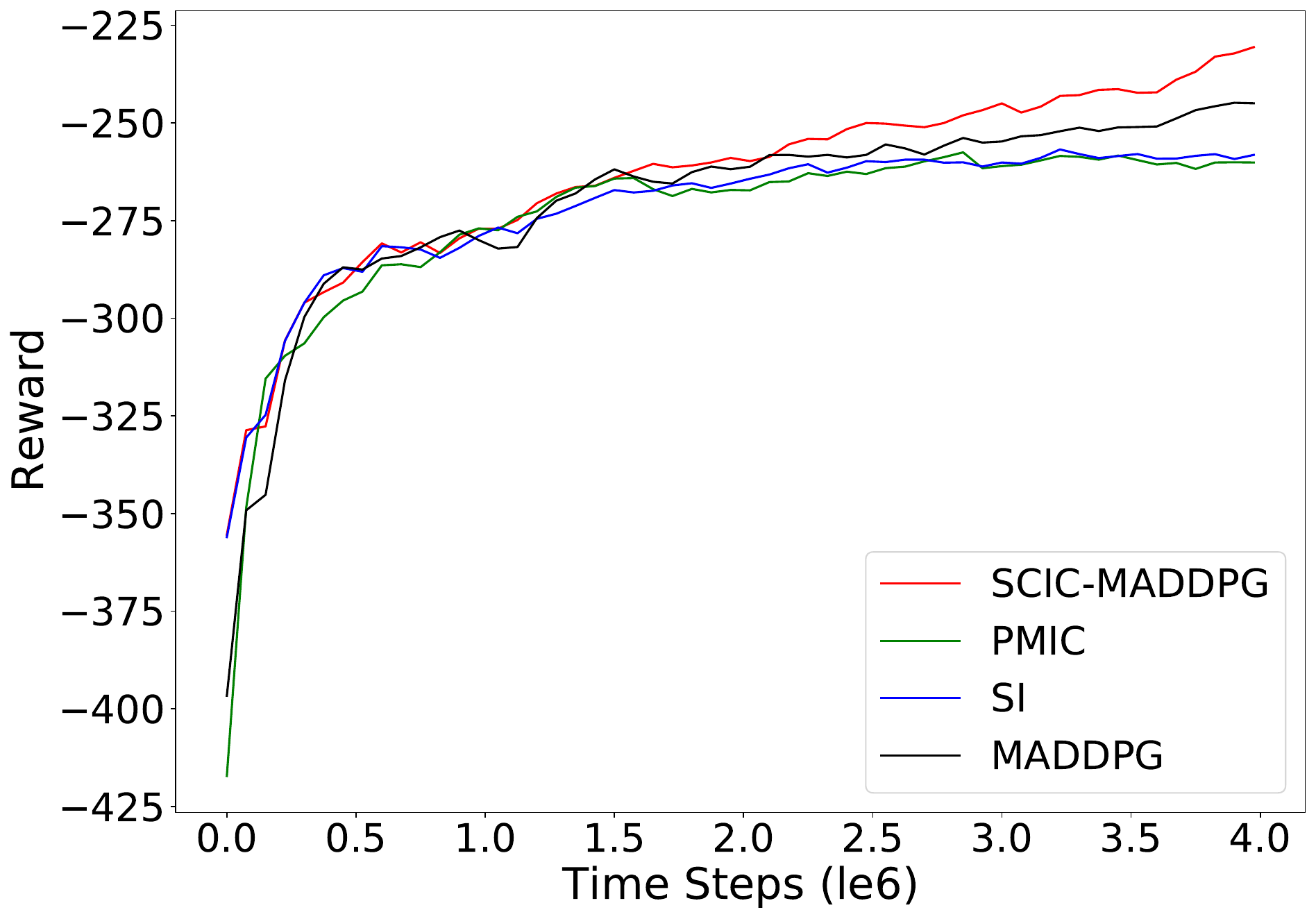}
	}
        \subfigure[Cooperative Navi. (5 agents).] {
	\centering
	\label{navigation5}
	\includegraphics[width=0.47\linewidth]{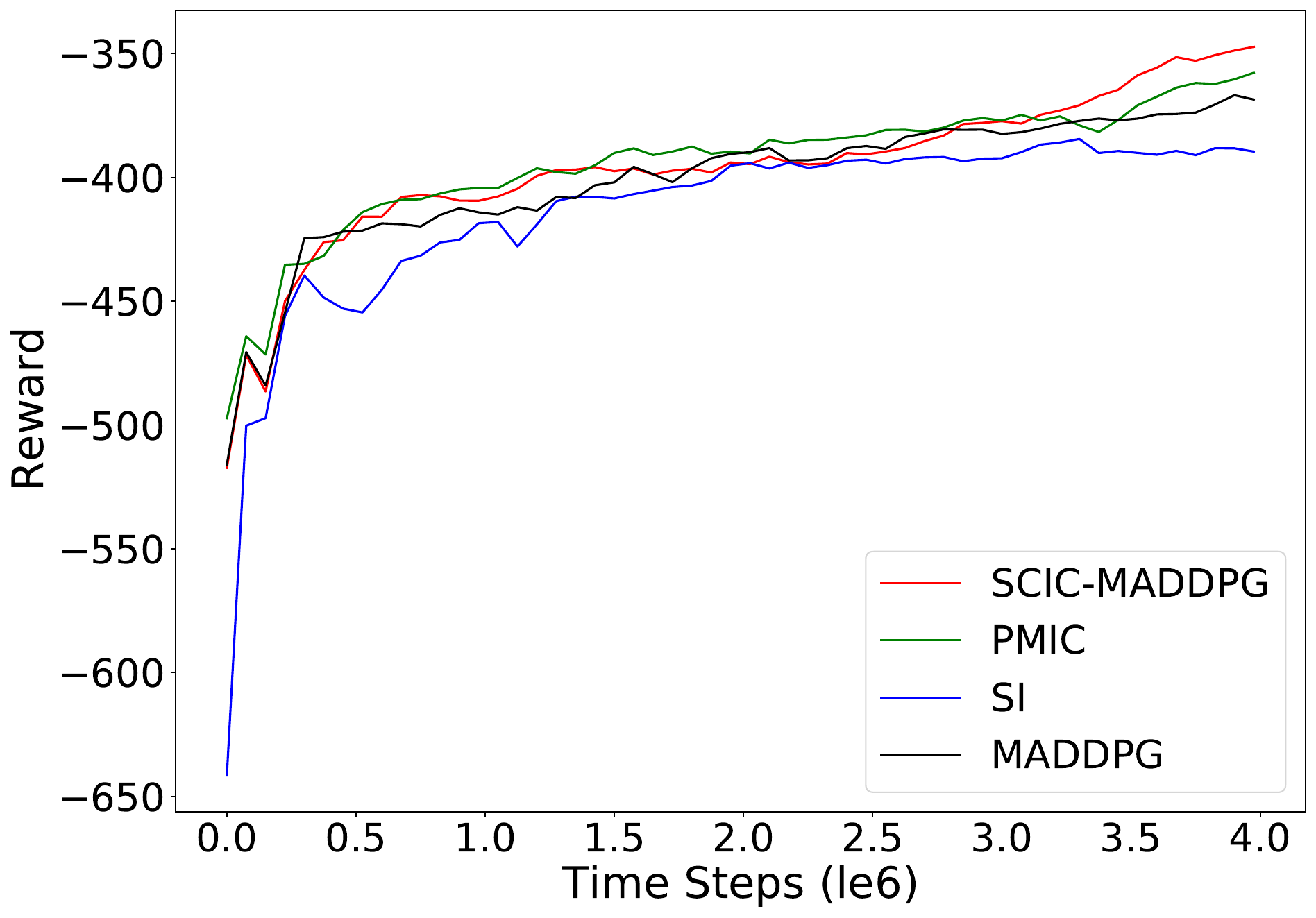}
	}
	\subfigure[Cooperative Line (3 agents).] {
	\centering
	\label{line3}
	\includegraphics[width=0.47\linewidth]{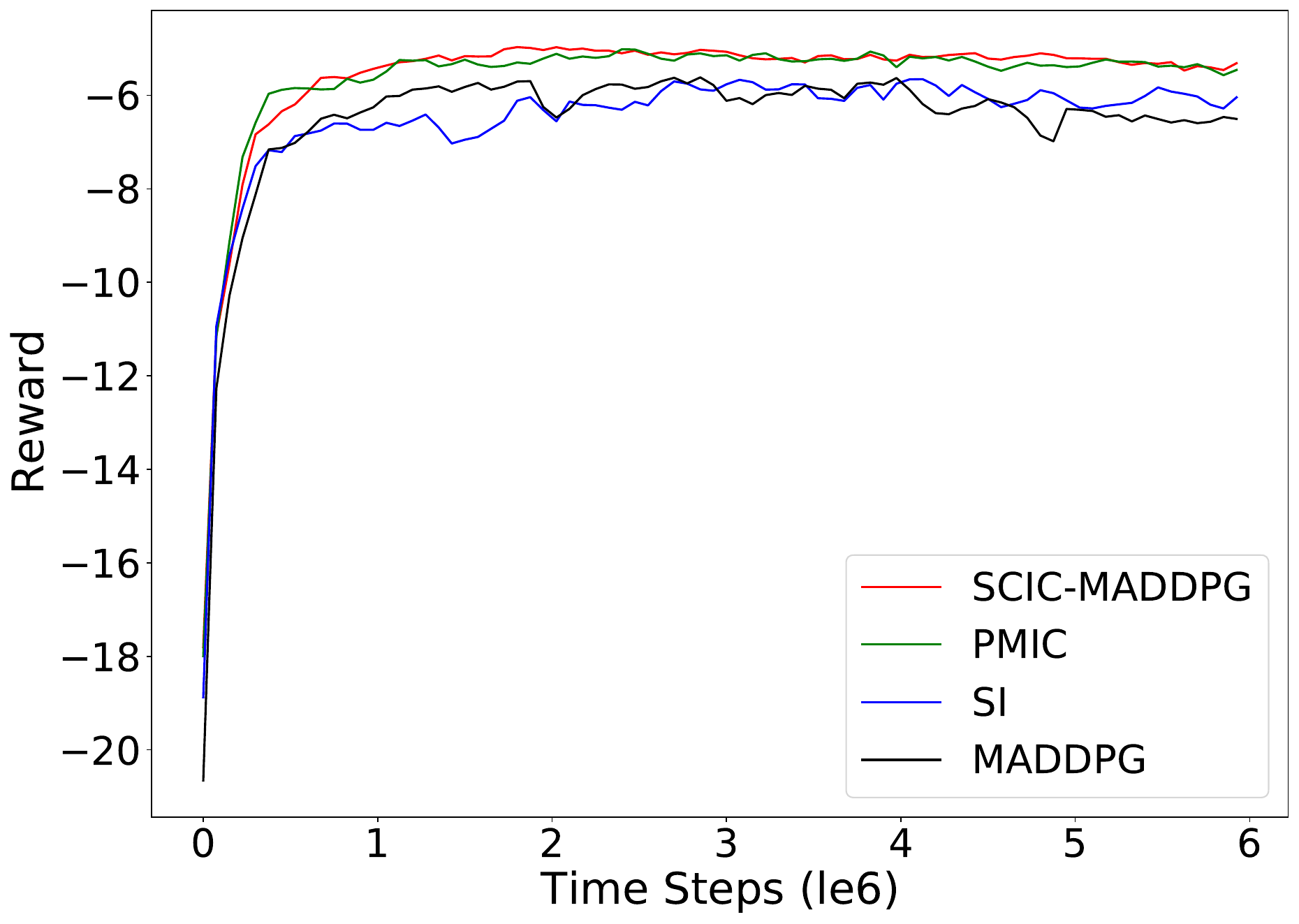}
	}
        \subfigure[Cooperative Line (5 agents).] {
	\centering
	\label{line5}
	\includegraphics[width=0.47\linewidth]{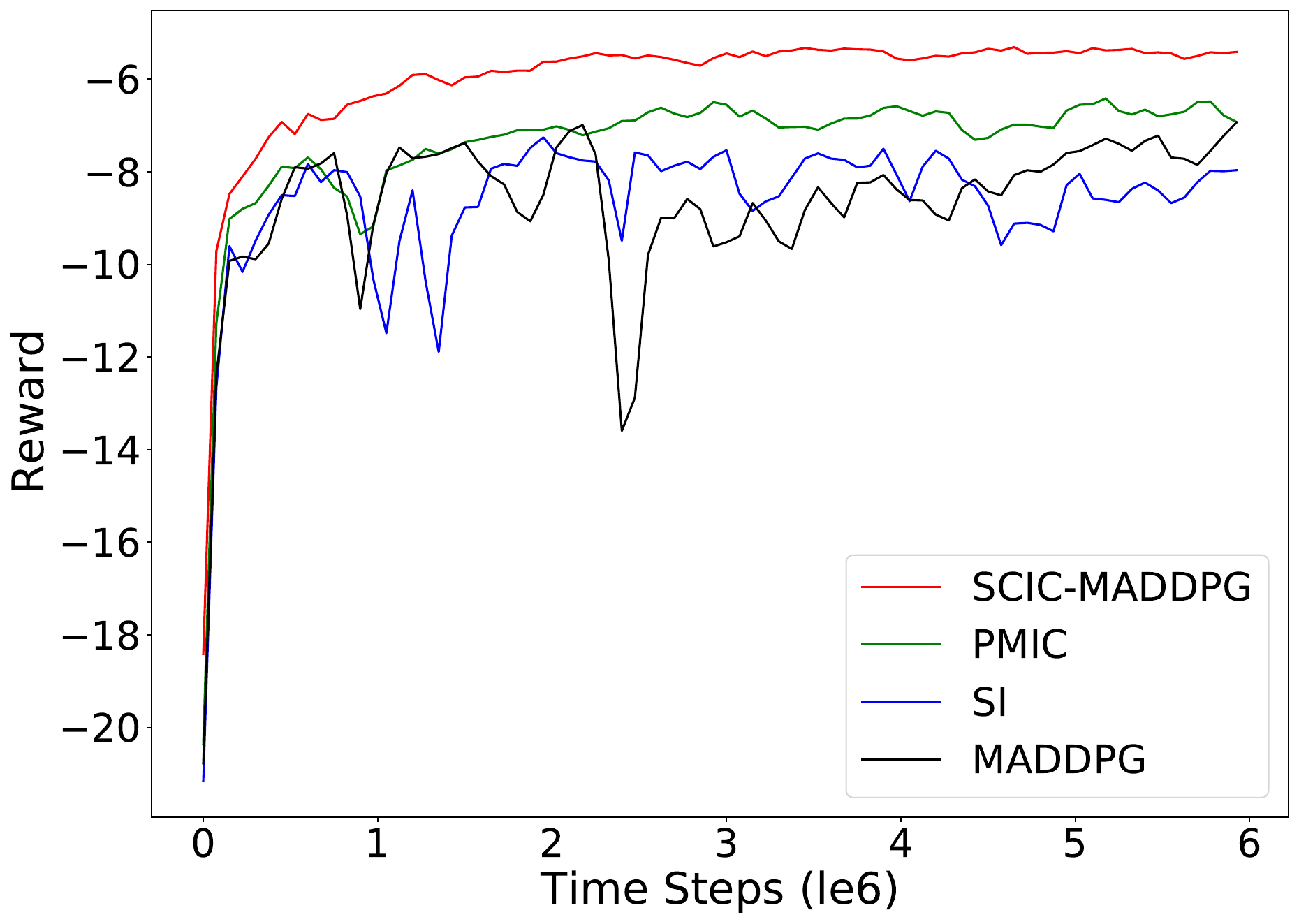}
	}

	\caption{Performances comparison of SCIC-MADDPG with the other three approaches in various multi-agent tasks. }
	\label{SCIC_evaluate}
\end{figure}
Since our approach is based on MADDPG with centralized training and decentralized executing, we name it SCIC-MADDPG in the experiments. 
First, we evaluate the performance of our approach and other baseline approaches on Cooperative Predator Prey with 3, 4, and 5 predators, in which the policy of predators needs to be trained and the policy of prey is fixed. Figure \ref{pre3}, Figure \ref{pre4}, and Figure \ref{pre5} illustrate the comparison results of rewards for SCIC-MADDPG, PMIC, SI, and MDDDPG on Predator-Prey tasks with 3, 4, 5 predators, respectively. We can observe that our SCIC-MADDPG demonstrates better performance compared to other algorithms, although there are relatively minor advantages when there are 4 predators in the environment. SCIC-MADDPG converges to a better reward than MADDPG, which indicates that the intrinsic reward is indeed conducive to improving the cooperation among agents. Both PMIC and SI receive lower rewards, 
suggesting that promoting agents to reach \textit{significant states} is more beneficial for cooperation between agents than merely focusing on behavior coordination.
From another perspective, enhancing the causal influence between agents is evidently more conducive to cooperation between agents than solely focusing on enhancing their correlation.

Moreover, we further perform evaluations on the Cooperative Navigation task with 3, 4, and 5 robots, which requires each agent to reach a distinct landmark to achieve the shortest total distance traveled. From Figure \ref{navigation3}, Figure \ref{navigation4}, and Figure \ref{navigation5}, we can observe that our SCIC-MADDPG consistently achieves higher rewards than the other baseline methods in the Cooperative Navigation tasks with varying numbers of agents. 
Furthermore, we evaluate the performance of SCIC-MADDPG on more challenging Control Line tasks involving 3 and 5 agents, representing an exceptionally difficult task scenario. The experimental findings illustrated in Figure \ref{line5} reveal that SCIC-MADDPG outperforms other methods in the context of a 5-agent scenario.  
Additionally, Figure \ref{line3} illustrates that SCIC-MADDPG achieves superior performance in comparison to SI and MADDPG, while achieving comparable performance to PMIC.
In summary, the results of these experiments collectively indicate that SCIC-MADDPG proves to be an effective approach,  consistently outperforming other methods across various tasks.

\begin{figure}[ht] \centering
	\subfigure[Ablation Performance.] {
	\centering
	\label{ablation_intervention}
	\includegraphics[width=0.47\linewidth]{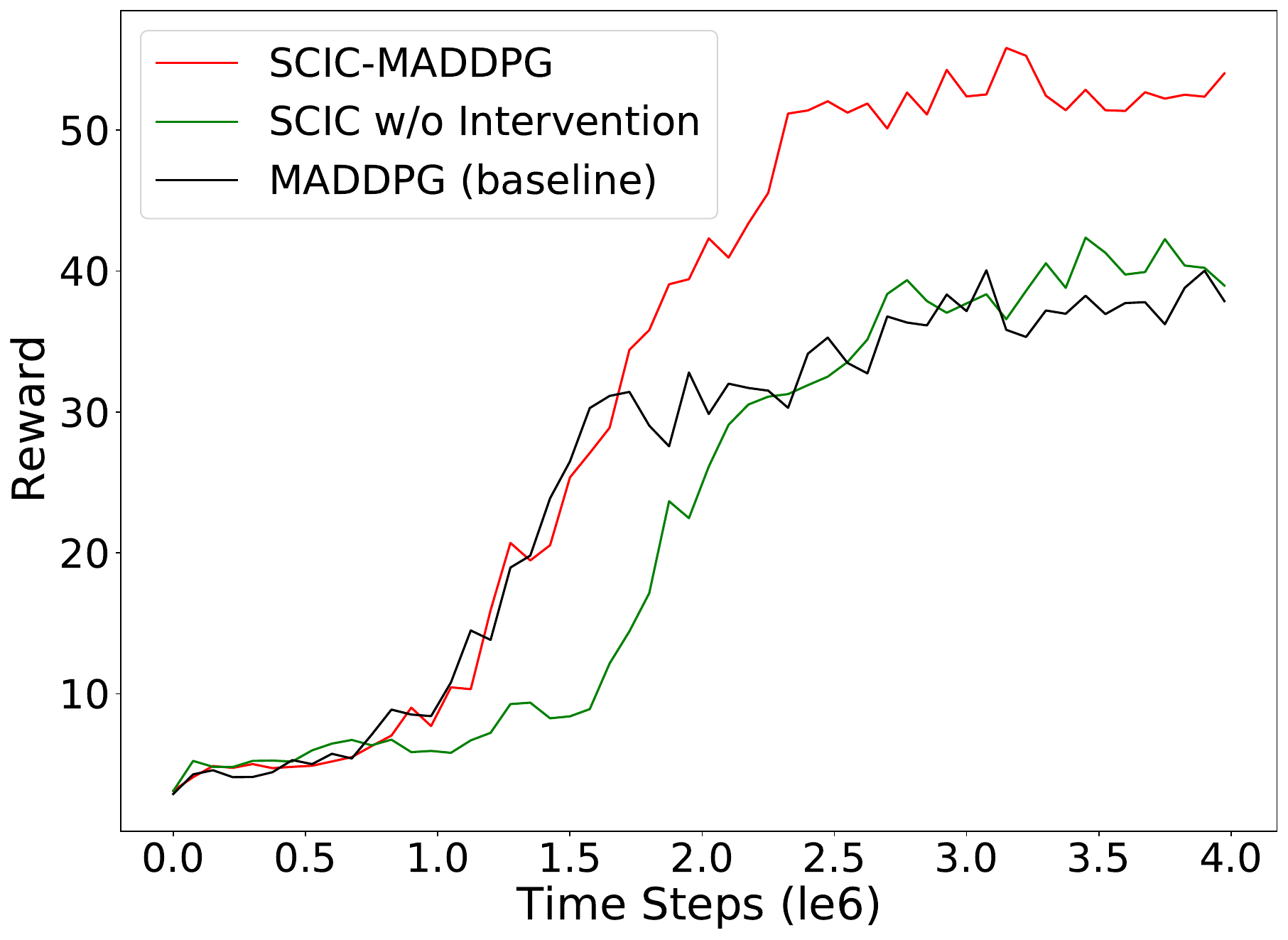}
	}
	\subfigure[Temperature Parameters.] {
	\centering
	\label{parameters}
	\includegraphics[width=0.47\linewidth]{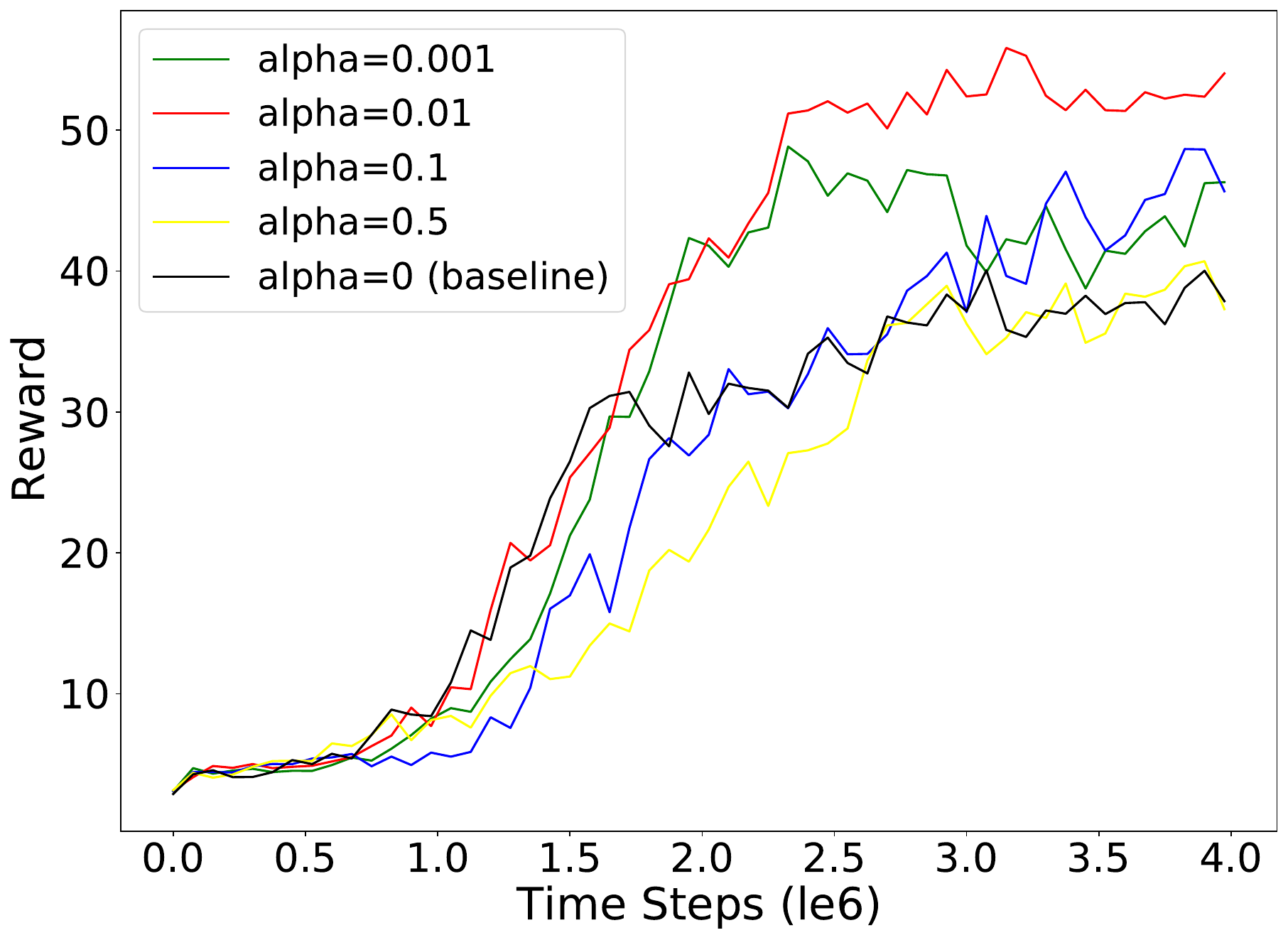}
	}
	\caption{Ablation Study.}
	\label{ablation}
\end{figure}

\subsection{Ablation study}
To further evaluate our proposed approach, we further conducted a series of ablation experiments in this subsection.  Specifically, to evaluate the effectiveness of intervention sampling, we implemented the SCIC w/o Intervention method, which obtains the action set when estimating causal effects between agents, not through intervention with a uniform distribution, but by sampling from the replay buffer. 
The ablation study was conducted on the Predator Prey task with 5 agents. As shown in Figure \ref{ablation_intervention},  SCIC-MADDPG performs significantly better than MADDPG and SCIC w/o Intervention. The key reason behind this achievement can be attributed to the fact that the sampling actions in the off-policy reinforcement learning algorithm originate from a mixture of different policies, which cannot be utilized to estimate causality.
The results reveal two facts: 1) Using causal influence between agents as a Reward Bonus contributes to enhancing the performance of MARL algorithms by facilitating cooperation among agents; 2) The adoption of intervention leads to more accurate estimates of causal influence.

\subsubsection{Temperature Parameters $\alpha$}
The role of temperature parameters $\alpha$ is to control the relative importance between intrinsic and extrinsic reward. We evaluate SCIC-MADDPG by varying $\alpha$ values within the range of [0, 0.001,0.01,0.1,0.5] in the Predator Prey task with 5 predators. As illustrated in Figure \ref{parameters}, SCIC-MADDPG with a temperature value of 0.01 outperforms its performance with other temperature values.

\section{Conclusion}\label{Conclusion}

To promote coordination between agents and encourage exploration, we propose the SCIC approach, which incorporates a new intrinsic reward mechanism based on a new cooperation criterion measured by situation-dependent causal influence between agents. SCIC encourages agents to explore states that positively affect other agents by detecting inter-agent causal influences and utilizing them as intrinsic rewards, thereby enhancing collaboration and overall performance. We conduct comprehensive experiments to evaluate the performance of our proposed approach across various cooperative MARL tasks. Extensive experimental results prove the effectiveness of our SCIC. In the future, we expect to further extend SCIC to decentralized training-based MARL algorithms and model-based MARL algorithms.

\clearpage

\section{Acknowledgments}
This work was supported by the National Key Research and Development Program of China (No. 2022ZD0119102). 

\bibliography{SCIC}

\clearpage

\subsection*{A \quad  Proof of Lemma 1}
First, the dependence $A_t^i \nupmodels S_{t+1}^j | S_t^i=s_t^i$ under an intervention $do($$A_t^i:=\pi(a_t^i|s_t^i))$ signifies that there are some $S_{t+1}^j$, and $a_t^{i,1}$, $a_t^{i,2}$ that satisfy the following conditions: 

\begin{align}
&p^{do(A_t^i:=\pi)}(s_{t+1}^j | s_t^i, a_t^{i,1}) = p(s_{t+1}^j | s_t^i, a_t^{i,1}) \nonumber
\\&\neq p(s_{t+1}^j | s_t^i, a_t^{i,2}) =p^{do(A_t^i:=\pi)}(s_{t+1}^j | s_t^i, a_t^{i,2}) \nonumber
\end{align}
Any $\pi^{\prime}$ with full support would satisfy $\pi^{\prime}(a_t^{i,1}|s_t^i) >0$ and $\pi^{\prime}(a_t^{i,2}|s_t^i) >0$. So the following formula holds, implying that the dependence  $A_t^i \nupmodels S_{t+1}^j | S_t^i=s_t^i$ holds under $do(A_t^i:=\pi^{\prime})$. 
\begin{align}
&p^{do(A_t^i:=\pi^{\prime})}(s_{t+1}^j | s_t^i, a_t^{i,1}) = p(s_{t+1}^j | s_t^i, a_t^{i,1}) \nonumber
\\&\neq p(s_{t+1}^j | s_t^i, a_t^{i,2})=p^{do(A_t^i:=\pi^{\prime})}(s_{t+1}^j | s_t^i, a_t^{i,2}) \nonumber
\end{align}
Since  $A_t^i \nupmodels S_{t+1}^j | S_t^i=s_t^i$ hold under all intervention with full support, the edge $A_t^i \rightarrow S_{t+1}^j$ exits, implying that $A_t^i$  has a causal effect on $S_{t+1}^j$. 

Second, if the independent $A_t^i \upmodels S_{t+1}^j | S_t^i=s_t^i$ hold under an intervention $do($$A_t^i:=\pi(a_t^i|s_t^i))$ with $\pi$ having full support, any intervention $do(A_t^i:=\pi^{\prime})$ holds that
\begin{align}
&p^{do(A_t^i:=\pi^{\prime})}(S_{t+1}^j | s_t^i, A_t^{i}) = p(S_{t+1}^j | s_t^i, A_t^{i}) \label{independent1}
\\& = p(S_{t+1}^j | s_t^i) = p^{do(A_t^i:=\pi^{\prime})}(S_{t+1}^j | s_t^i), \label{independent2}
\end{align}
where the equation \ref{independent1} contributes to the autonomy property of causal mechanisms, the equation \ref{independent2} is due to the independence $A_t^i \upmodels S_{t+1}^j | S_t^i=s_t^i$. So, if $A_t^i \upmodels S_{t+1}^j$ holds under an intervention $do($$A_t^i:=\pi(a_t^i|s_t^i))$ with $\pi$ having full support, then the independence holds. So far, Lemma 1 has been proven.

\end{document}